\documentclass[conference]{IEEEtran}
\IEEEoverridecommandlockouts
\usepackage{cite}
\usepackage{amsmath,amssymb,amsfonts}
\usepackage{algorithmic}
\usepackage{graphicx}
\usepackage{textcomp}
\usepackage{xcolor}
\usepackage{pdfpages}
\usepackage{stfloats}
\usepackage{subcaption}
\usepackage{url}
\usepackage{tabularx}
\usepackage{algorithm}
\usepackage{algorithmic}

\def\BibTeX{{\rm B\kern-.05em{\sc i\kern-.025em b}\kern-.08em
    T\kern-.1667em\lower.7ex\hbox{E}\kern-.125emX}}
\begin{document}

\title{
FilterMobileViT  and DropoutVIT}

\author{\IEEEauthorblockN{1\textsuperscript{st} Bohang Sun}
\IEEEauthorblockA{\textit{School of Information and Software Engineering} \\
\textit{University of Electronic Science and Technology of China}\\
Chengdu, China \\
bobsun@std.uestc.edu.cn}

\and
\IEEEauthorblockN{2\textsuperscript{st} Bo Chen}
\IEEEauthorblockA{\textit{School of Information and Software Engineering} \\
\textit{University of Electronic Science and Technology of China}\\
Chengdu, China \\
bochen@uestc.edu.cn}
}
\maketitle

\begin{abstract}
In this study, we propose an improved variant of ViT that performs attention-based QKV operations in the early stages of downsampling. Directly conducting attention operations on high-resolution feature maps is computationally intensive due to the large size and numerous tokens. To address this issue, we introduce a filter attention mechanism using Filter Block to generate an salient mask (Filter Mask) that selects the most informative pixels for attention computation.

Specifically, the Filter Block scores the pixels of the feature map, and we sort these scores to select the top $K$ pixels (with $K$ varying across network layers). This method effectively reduces the number of tokens involved in the attention computation, thereby lowering computational complexity and increasing processing speed. Additionally, we find that the salient mask offers interpretability, as the model focuses more on regions of the image that are crucial to the outcome.

Experimental results demonstrate that our model achieves advantages in parameter efficiency and computational speed while improving accuracy. Compared to other models, our approach significantly reduces computational resource consumption while maintaining high performance.
\end{abstract}

\begin{IEEEkeywords}
MobileViT, Transformer, Efficient Attention,Learned Sparse Attention, Convolution,Interpretability
\end{IEEEkeywords}

\section{Introduction}
Vision Transformer (ViT) \cite{dosovitskiy2020image} has revolutionized the field of computer vision by introducing transformer-based architectures to image tasks. ViT leverages attention mechanisms over 16×16 image patches, allowing for global context learning across the image. However, while ViT achieves state-of-the-art results on various tasks, its reliance on large tokens makes it computationally expensive when applied to high-resolution images. The quadratic complexity of the QKV operation \cite{vaswani2017attention} means that increasing the number of tokens dramatically increases the computational burden.

To address this issue, MobileViT \cite{mehta2021mobilevit} was introduced as a lighter version of ViT, which performs attention after downsampling the image. This reduces the number of tokens involved in attention, thus improving computational efficiency. However, MobileViT sacrifices granularity by operating on downsampled feature maps (e.g., 56×56 or 28×28), which may limit its ability to capture finer details. Moreover, not all pixels contribute equally to the final prediction. Many pixels are noisy or irrelevant, while others are critical for decision-making.

In this paper, we propose a new variant of MobileViT, called \textbf{FilterMobileViT }, that allows for more granular attention without the need for early downsampling. Our approach leverages a convolutional neural network (CNN) to generate an salient mask, which determines which pixels in the feature map are most relevant for attention. By sorting the pixel salient scores and selecting the top $K$ pixels, we perform attention on only the most significant parts of the image, thereby reducing the computational complexity of the QKV operation.

Unlike the challenging task of implementing sparse matrix operations on GPUs, our method offers a practical solution by selecting salient pixels directly. This ensures that the computational load is reduced while maintaining accuracy. Additionally, our approach offers interpretability, as the salient mask highlights the critical areas of the image that the model focuses on during attention. For example, in an image of a lamb, the mask effectively highlights the contour of the lamb while ignoring irrelevant background features.

In summary, our contributions are threefold:
\begin{itemize}
    \item We introduce a lightweight, more efficient attention mechanism that allows for finer-grained attention while reducing computational complexity.
    \item Our method inherently provides interpretability by focusing attention on the most relevant regions of the image, as demonstrated through visualizations of the salient mask.
    \item Through an ablation study, we introduce \textbf{DropoutVIT}, a variation of our model that mimics dropout by randomly selecting pixels, further validating the flexibility and robustness of our approach.
\end{itemize}

\section{Related Work}

\subsection{Vision Transformers}
Vision Transformers (ViT) \cite{dosovitskiy2020image} have demonstrated significant success in computer vision tasks by applying attention mechanisms to image patches, typically 16×16 patches. While ViT excels at capturing global dependencies across the image, the quadratic complexity of its QKV operations \cite{vaswani2017attention} poses a computational challenge, particularly for high-resolution images. The number of tokens increases rapidly, leading to significantly higher computational costs as the image resolution grows.

\subsection{MobileViT}
MobileViT \cite{mehta2021mobilevit} addresses this issue by downsampling the image before applying transformer-based attention. By performing attention at smaller resolutions (e.g., 56×56 or 28×28), it reduces the number of tokens and thus lowers computational complexity. However, this downsampling sacrifices the granularity of attention, as finer details may be missed when attention is applied to a lower-resolution feature map.

\subsection{Efficient Attention Mechanisms}
Several attention-based models have been proposed to optimize computational complexity. Longformer \cite{beltagy2020longformer} employs a sparse attention mechanism, restricting attention to a local window around each token, which reduces the computational load but may struggle to capture global dependencies in images. Performer \cite{choromanski2020rethinking} approximates softmax attention using kernel-based methods, achieving linear complexity with respect to sequence length, a promising approach for reducing the cost of attention in vision tasks. Linformer \cite{wang2020linformer} further reduces complexity by projecting keys and values into lower dimensions, maintaining linear complexity while conserving memory. Reformer \cite{reformer} introduces locality-sensitive hashing (LSH) to approximate nearest neighbors in attention space, making the computation more efficient, though this approach may face challenges when highly detailed structures require attention.

\subsection{Lightweight Convolutional Architectures}
In addition to attention mechanisms, lightweight convolutional architectures have been developed to balance efficiency and performance. MobileNetV2 \cite{sandler2018mobilenetv2} introduces an inverted residual block to optimize information flow with fewer parameters. MobileNetV3 \cite{howard2019searching} improves on this by using neural architecture search (NAS) to design an efficient structure, optimizing both latency and accuracy. EfficientNet \cite{tan2019efficientnet} scales networks by balancing depth, width, and resolution, achieving optimal performance across configurations. GhostNet \cite{han2020ghostnet} reduces feature map redundancy by generating fewer feature maps, thus improving speed and reducing parameter count. Finally, LCNet \cite{cui2021pp} focuses on low-complexity convolutional operations to achieve faster inference times while maintaining reasonable accuracy, making it ideal for edge computing tasks.

\subsection{Filter Attention Mechanism}
Our work builds upon these advancements by introducing the Filter Attention mechanism. Unlike models that downsample or linearize attention computations, our approach selectively applies attention to the most salient pixels using a CNN-generated filter mask. By sorting the pixels based on their salient and selecting the top $K$ pixels, we reduce the number of tokens involved in the QKV computation. This maintains global context while significantly reducing computational complexity, without sacrificing attention granularity. Our approach also offers interpretability, as visualizing the filter mask shows that the model effectively focuses on the most relevant parts of the image.

\section{Method}
\subsection{Structure}
In our proposed network architecture, we combine convolutional layers with transformer layers to balance local feature extraction and global attention. The key innovation of our method is the introduction of a Filter Attention mechanism, which selectively attends to salient regions of the feature map, significantly reducing the computational complexity of attention-based operations.

The network processes the input image through a CNN, which produces a feature map. This feature map is treated as a set of tokens that are reshaped and passed through a Transformer Encoder \cite{vaswani2017attention}. Unlike traditional attention mechanisms, our Filter Attention selectively identifies salient tokens based on a filter mask generated by the CNN, which is applied to reduce the number of tokens used in the attention computation. This mechanism allows the model to ignore irrelevant regions of the image, as not every pixel contributes equally to the final prediction. In theory, some pixels may represent noise or unnecessary details, and even among useful pixels, their contributions to the prediction are not uniform \cite{xie2022vit}.

\begin{figure*}[htbp]
    \centering
    \includegraphics[width=0.8\textwidth]{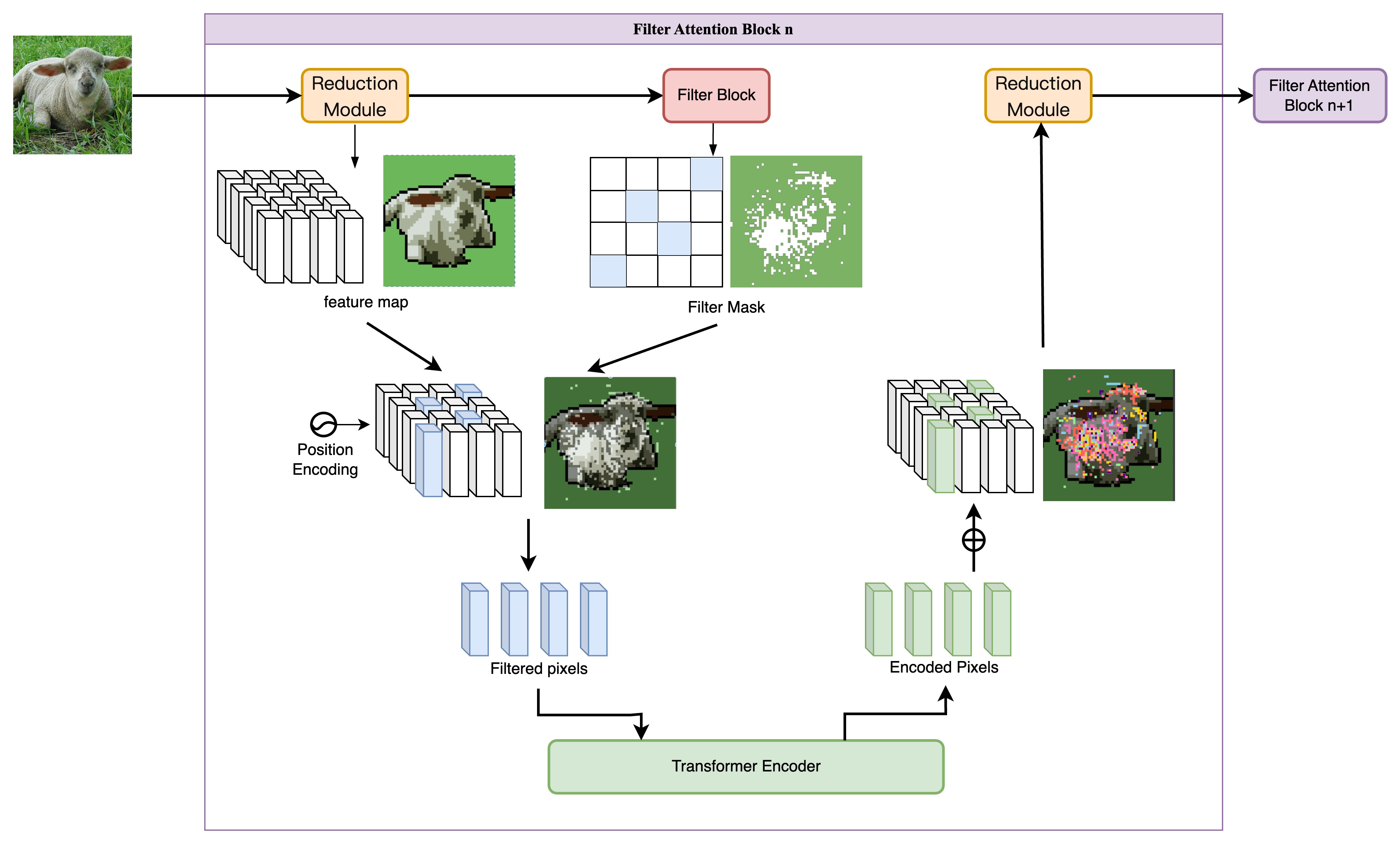}
    \caption{An overview of the Filter Attention mechanism. The input image is first processed by a convolutional neural network (CNN) to generate a feature map. A filter mask is then applied to this feature map, which selects the most salient pixels for further attention-based computation. Only the top-ranked pixels, based on the salient scores, are passed through the Transformer Encoder for global attention processing. This reduces the computational complexity by focusing attention on the most relevant regions of the image.}
    \label{fig:filter_attention}
\end{figure*}

\subsection{Filter Attention Block}

The core of our method is the Filter Attention block, which integrates a convolutional block with a transformer encoder. Instead of processing all tokens, we use a filter mask to select only the most salient ones for attention computation, significantly reducing complexity.

Initially, an image is processed through a CNN module, generating a feature map \( X_{f} \in \mathbb{R}^{N \times C \times H \times W} \) and a filter mask \( M \in \mathbb{R}^{N \times 1 \times H \times W} \) that assigns saliency scores to each pixel. The feature map is then element-wise multiplied by the mask, retaining only the most relevant pixels:

\[
X_{\text{filtered}} = X_{f} \odot M
\]

Next, \( X_{\text{filtered}} \) is reshaped into a transformer-compatible form \( X_{t} \in \mathbb{R}^{N \times K \times C} \), where \( K \ll H \times W \) represents the selected tokens. The filtered tokens are passed through a Transformer Encoder~\cite{vaswani2017attention}, yielding the output \( X_{t}' \).

\subsubsection{Computational Complexity Reduction}

In the original attention mechanism, the computation complexity is \( O(T^2 d_k) \), where \( T = H \times W \) represents the number of tokens. By reducing the tokens to \( K \), the attention computation simplifies to:

\[
\text{Attention}(\mathbf{Q}', \mathbf{K}', \mathbf{V}') = \text{softmax}\left( \frac{\mathbf{Q}' (\mathbf{K}')^\top}{\sqrt{d_k}} \right) \mathbf{V}',
\]

with complexity \( O(K^2 d_k) \), achieving substantial computational savings when \( K \ll T \).

This reduction allows efficient attention on higher-resolution features, enabling finer-grained attention without excessive computational cost. The Filter Attention mechanism is detailed in Algorithm~\ref{alg:filterattention}, focusing on informative pixels and preserving critical spatial information.

\begin{algorithm}[htbp]
\caption{FilterAttention Block}
\label{alg:filterattention}
\begin{algorithmic}[1]
\REQUIRE Feature map \( x \in \mathbb{R}^{B \times C \times H \times W} \)
\STATE Compute saliency map: \( \text{imp} = \sigma(\text{Conv}(x)) \)
\STATE Select top \( K \) indices per sample: \( \text{indices} = \text{TopK}(\text{imp}, K) \)
\STATE Mask feature map: \( x \leftarrow x \odot \text{imp} \)
\STATE Extract selected pixels and add positional encoding:
\[
\text{sel\_pixels}[b, k] = x[b, :, i, j] + \text{pos}[i \times W + j]
\]
\hspace{1em} where \( (b, i, j) \in \text{indices} \)
\STATE Apply Transformer Encoder:
\[
\text{tokens} = \text{TransformerEncoder}(\text{sel\_pixels})
\]
\STATE Update feature map at selected indices:
\[
x[b, :, i, j] \leftarrow \text{tokens}[b, k]
\]
\hspace{1em} where \( (b, i, j) \in \text{indices} \)
\RETURN Updated feature map \( x \)
\end{algorithmic}
\end{algorithm}

\subsection{Inverted Residual Block}
In our network, we employ the Inverted Residual Block \cite{sandler2018mobilenetv2} to enhance feature extraction while maintaining model efficiency. This block is designed to first expand the input channels, apply depthwise convolution, and then project the output back to the original dimensions, creating a residual connection. This structure ensures that important spatial information is preserved while reducing the number of parameters in the model, making it computationally efficient for both small and large-scale vision tasks.

\begin{figure*}[h]
    \centering
    \includegraphics[width=0.7\textwidth]{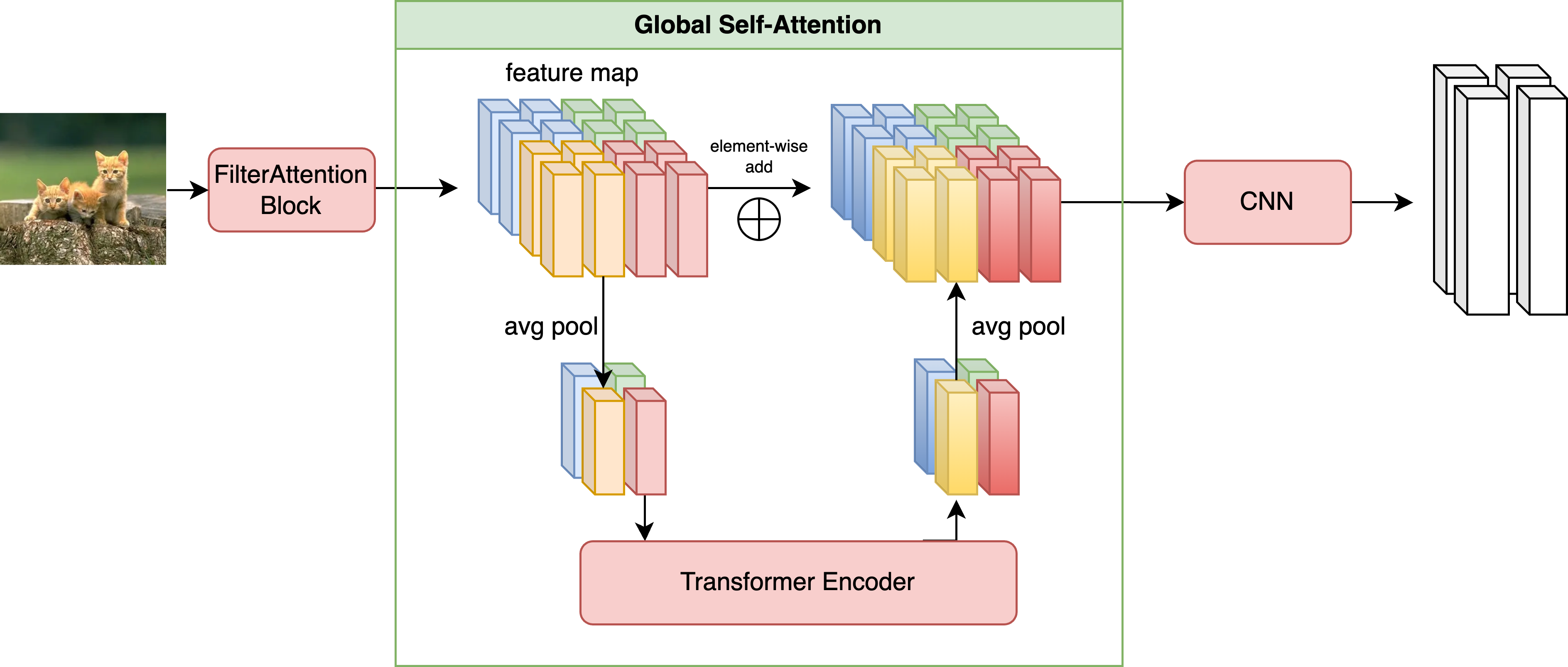}
    \caption{Illustration of the Global Self-Attention mechanism with pooling. After the initial feature extraction by the CNN, average pooling is applied to reduce the size of the feature map. This pooling operation significantly decreases the number of tokens passed into the Transformer Encoder for self-attention computation. The remaining tokens retain global context and are processed by the self-attention module, capturing long-range dependencies in the image with reduced computational cost.}
    \label{fig:global_attention}
\end{figure*}

\subsection{Global Self-Attention with Pooling}
To further reduce the computational complexity of the attention mechanism, we apply average pooling to the feature map before feeding it into the self-attention layers. Pooling reduces the spatial dimensions of the input feature map, which lowers the number of tokens passed into the attention module, thereby decreasing the quadratic cost of QKV operations. Despite this reduction, global context is maintained, ensuring that the model can still capture long-range dependencies between the most relevant regions of the image.

\subsection{Transformer Encoder}
The selected tokens from the filtered feature map are processed through the Transformer Encoder \cite{vaswani2017attention}. The transformer layer applies self-attention to capture global dependencies between tokens. After passing through the transformer, the tokens are reshaped back into their original spatial dimensions and processed through subsequent CNN layers for further refinement.

Finally, the output feature map is used for classification or other downstream tasks, leveraging both local and global information extracted from the image.

\section{Experiment}

We conducted experiments using five distinct img-100 subsets, each randomly sampled from the ImageNet-1K dataset \cite{russakovsky2015imagenet}. The results presented in this paper are based on the first img-100 subset. To ensure a fair comparison, the same training settings and hyperparameters were applied uniformly across all models. The specific class selections for the five img-100 subsets used in this study are provided on GitHub for reference and reproducibility at the following address: \url{https://github.com/BobSun98/FilterVIT}.

\subsection{Dataset}

The img-100 subsets were constructed from 100 randomly selected classes within the ImageNet-1K dataset \cite{russakovsky2015imagenet}. Training and validation were performed on these subsets. Each model was trained on five distinct subsets to enhance robustness and reduce overfitting, ensuring that the reported results are not skewed by the selection of a particular subset. The smaller img-100 dataset allows for faster prototyping and algorithm validation while still posing a challenge in reducing overfitting. If lightweight models perform well on a smaller dataset, it indicates their potential for better generalization.

\subsubsection{Data Augmentation}

To improve the generalization ability of all models, we applied consistent data augmentation techniques during training. These techniques included RandomResizedCrop(224) to randomly crop and resize images to \(224 \times 224\) pixels, RandomHorizontalFlip to randomly flip images horizontally, TrivialAugmentWide \cite{muller2021trivialaugment} to introduce broad augmentation variability, and normalization with a mean of \([0.5, 0.5, 0.5]\) and a standard deviation of \([0.5, 0.5, 0.5]\). During validation, images were resized to \(256 \times 256\) pixels, followed by a center crop to \(224 \times 224\) pixels, with normalization applied using the same mean and standard deviation values.

\subsection{Models and Baselines}

To evaluate the effectiveness of our proposed \textbf{FilterMobileViT}, we compared it against several baseline models that represent a range of architectures in the field of efficient neural networks. These include \textbf{MobileNetV2} \cite{sandler2018mobilenetv2}, known for its inverted residual blocks and linear bottlenecks, and \textbf{MobileNetV3} \cite{howard2019searching}, which improves upon its predecessor with neural architecture search and hard-swish activation functions. We also included \textbf{EfficientNet-Lite0} \cite{tan2019efficientnet}, which balances network depth, width, and resolution for optimal efficiency, and \textbf{GhostNet} \cite{han2020ghostnet}, which reduces redundancy in feature maps to improve computational performance.

In addition to convolutional architectures, we compared against lightweight transformer-based models such as \textbf{Tiny-ViT} \cite{tinyvit2022}, a compact vision transformer designed for high performance with fewer parameters, and \textbf{LeViT} \cite{graham2021levit}, a hybrid model that combines convolutional and transformer layers for efficient inference. We also included \textbf{TinyNet} \cite{han2020model}, a family of models optimized for small parameter sizes and computational costs, and \textbf{LCNet} \cite{cui2021pp}, which prioritizes speed in its lightweight convolutional network design. Finally, \textbf{MobileViT-S} \cite{mehta2021mobilevit} was included as it incorporates transformer layers into a mobile-friendly architecture, similar to our approach.

These models were selected to provide a comprehensive comparison across different architectural strategies, highlighting the strengths of our \textbf{FilterMobileViT} in terms of accuracy and efficiency.

\subsection{Training Details}

All models were trained with consistent hyperparameters: a batch size of 64 and the AdamW optimizer \cite{loshchilov2017decoupled} (\(\beta_1 = 0.9\), \(\beta_2 = 0.999\), and a weight decay of 0.01). The initial learning rate was set to 0.0005 and decayed following a cosine annealing schedule \cite{loshchilov2016sgdr} to a minimum of \(1 \times 10^{-5}\), over 120 epochs. CosineAnnealingLR was used as the learning rate scheduler, with \(T_{\text{max}} = 120\) epochs. Training was performed on a Tesla P40 GPU, while inference performance was measured on both an RTX 4090 GPU for CUDA and an Apple M1 Pro chip for CPU. These configurations allowed for a comprehensive comparison of model efficiency across different hardware environments, ranging from high-performance GPUs to resource-constrained CPUs.


\subsection{Experimental Results}

As shown in Table~\ref{tab:results}, \textbf{FilterMobileViT} achieves a strong balance between accuracy and computational efficiency. With only 1.89 million parameters, it achieves an accuracy of 0.861, outperforming most other models except for \textbf{Tiny-ViT} \cite{tinyvit2022}, which achieves a slightly higher accuracy of 0.876 but with significantly more parameters (10.59M). This indicates that while Tiny-ViT offers marginally better performance, FilterMobileViT provides a more efficient solution in terms of model size, making it more suitable for deployment in resource-constrained environments.

FilterMobileViT also demonstrates competitive performance in terms of frames per second (FPS). Although it falls slightly behind models like \textbf{LCNet\_100}, which achieves higher FPS on both CPU and CUDA platforms, LCNet prioritizes speed over accuracy, resulting in a lower accuracy of 0.808. In contrast, FilterMobileViT offers a better trade-off between speed and accuracy, maintaining high accuracy with reasonable computational efficiency.

Moreover, compared to other lightweight models such as \textbf{MobileNetV2} \cite{sandler2018mobilenetv2} and \textbf{GhostNet} \cite{han2020ghostnet}, FilterMobileViT not only reduces the parameter count but also enhances accuracy. For instance, MobileNetV2 has 2.35 million parameters with an accuracy of 0.833, while GhostNet has 4.02 million parameters and achieves an accuracy of 0.842. This improvement underscores the effectiveness of our Filter Attention mechanism in selectively focusing on the most informative regions of the image, leading to better performance without increasing computational costs.

In terms of inference speed, while \textbf{EfficientNet\_Lite0} and \textbf{MobileNetV3\_Small} offer higher FPS on CUDA, they suffer from lower accuracy compared to FilterMobileViT. This further highlights the advantage of our approach in achieving a balanced performance suitable for practical applications where both accuracy and efficiency are critical.

Overall, FilterMobileViT demonstrates that incorporating the Filter Attention mechanism can lead to significant improvements in accuracy and efficiency, outperforming several established models in the field. Its lower parameter count and competitive speed make it an attractive option for deployment on devices with limited computational resources.

\begin{table*}[htbp]
    \centering
    \caption{Comparison of Models on the First Subset}
    \begin{tabular}{|l|c|c|c|c|}
    \hline
    \textbf{Model} & \textbf{\#Params (M)} & \textbf{FPS (CPU)} & \textbf{FPS (CUDA)} & \textbf{Accuracy} \\
    \hline
    \textbf{MobileNetV2}            & 2.35             & 11.29         & 848            & 0.833        \\
    \textbf{MobileNetV3\_Large}     & 4.33             & 6.57          & 742            & 0.832        \\
    \textbf{MobileNetV3\_Small}     & 1.62             & 7.14          & 892            & 0.786        \\
    \textbf{EfficientNet}         & 3.49             & 3.63          & 873            & 0.841        \\
    \textbf{Tiny-ViT}            & 10.59            & 6.19          & 524            & \textbf{0.876}    \\
    \textbf{LeViT}                  & 8.52             & 12.96         & 442            & 0.787        \\
    \textbf{GhostNet}               & 4.02             & 8.60          & 464            & 0.842        \\
    \textbf{TinyNet}                  & 5.03             & 2.75          & 475            & 0.848        \\
    \textbf{LCNet}                  & 1.80             & \textbf{7.29}      & \textbf{1377}       & 0.808        \\
    \textbf{MobileViT}                & 5.00             & 6.14          & 485            & 0.851        \\
    \textbf{FilterMobileViT (ours)}      & \textbf{1.89}         & 5.60          & 410            & \textbf{0.861}    \\
    \hline
    \end{tabular}
    \label{tab:results}
\end{table*}

\subsection{Accuracy and Loss Curves}

\begin{figure}[htbp]
    \centering
    \includegraphics[width=0.45\textwidth]{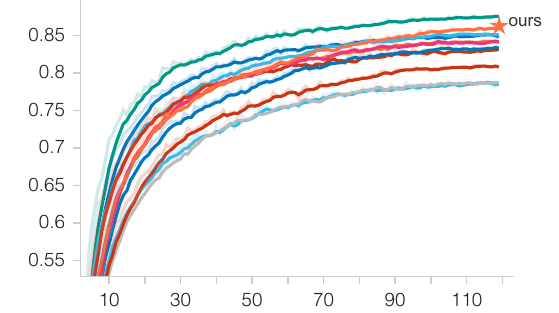}
    \caption{Validation accuracy for the first img-100 subset.}
    \label{fig:accuracy_img100}
\end{figure}

\begin{figure}[htbp]
    \centering
    \includegraphics[width=0.5\textwidth]{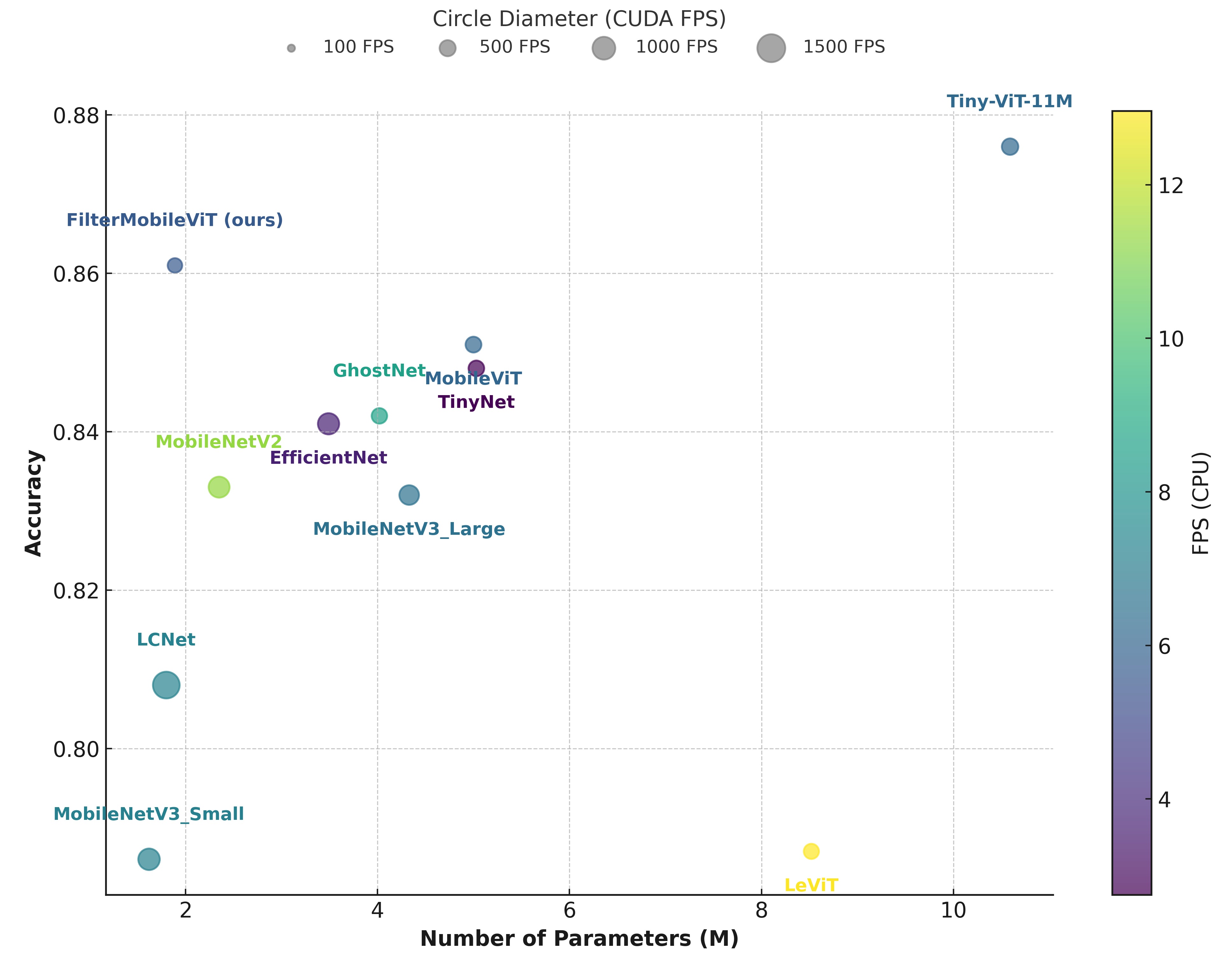}
    \caption{Scatter plot comparing different models in terms of parameter count, accuracy(for first imageNet-100 subset), and inference speed. The x-axis represents the number of parameters (in millions), and the y-axis represents classification accuracy. Each point’s color indicates the inference speed on the CPU (FPS), with a gradient from purple to yellow representing increasing speed. The diameter of each point corresponds to the inference speed on CUDA (FPS), as indicated by the horizontal legend at the top (showing example values of 100, 500, 1000, and 1500 FPS).}
    \label{fig:modelsScatterplot}
\end{figure}


Figures~\ref{fig:accuracy_img100} shows the validation accuracy for the first img-100 subset. As depicted in the accuracy curve, FilterMobileViT converges rapidly and maintains competitive accuracy throughout the training process, closely approaching Tiny-ViT's performance while maintaining a much smaller parameter count. The results for the other four img-100 subsets can be found in the Appendix.

\section*{Interpretability}
Previous research has aimed to explain and visualize the behavior of vision models and Vision Transformers (ViT). For instance, \cite{hendricks2021attention} introduced the Attention Rollout method to visualize attention maps, providing insights into where the model focuses during prediction. Similarly, \cite{kumar2021visualizing} developed techniques to analyze the internal representations of ViTs by clustering their learned features. Furthermore, Vit-CX \cite{xie2022vit} leverages representations from the last layer of ViTs and employs clustering to represent the model's focus areas. In our experiments, we found that FilterMobileViT  inherently possesses interpretability without the need for additional visualization techniques.

To better understand the behavior of FilterMobileVit, we visualized the filter masks generated by each Filter Attention layer. This was done by running inference on a set of test images from a pre-trained model and extracting the filter masks at different layers. These masks were then resized to match the original image dimensions and overlaid on the input images to provide a clear visual representation of where the model is focusing its attention \cite{selvaraju2017grad}.

In the first Filter Attention layer, the model tends to focus on both the foreground and background, sometimes highlighting object edges. This suggests that the first layer is responsible for identifying broad and general features across the image. As we move deeper into the second layer, the focus shifts more towards the main object or foreground of the image, indicating that the model has started to refine its attention and concentrate on the most salient features. In the third and final layer, the attention shifts back to the background, possibly to refine the overall context and ensure that irrelevant regions are appropriately weighted.

\begin{figure}[htbp]
    \centering
    \includegraphics[width=0.5\textwidth]{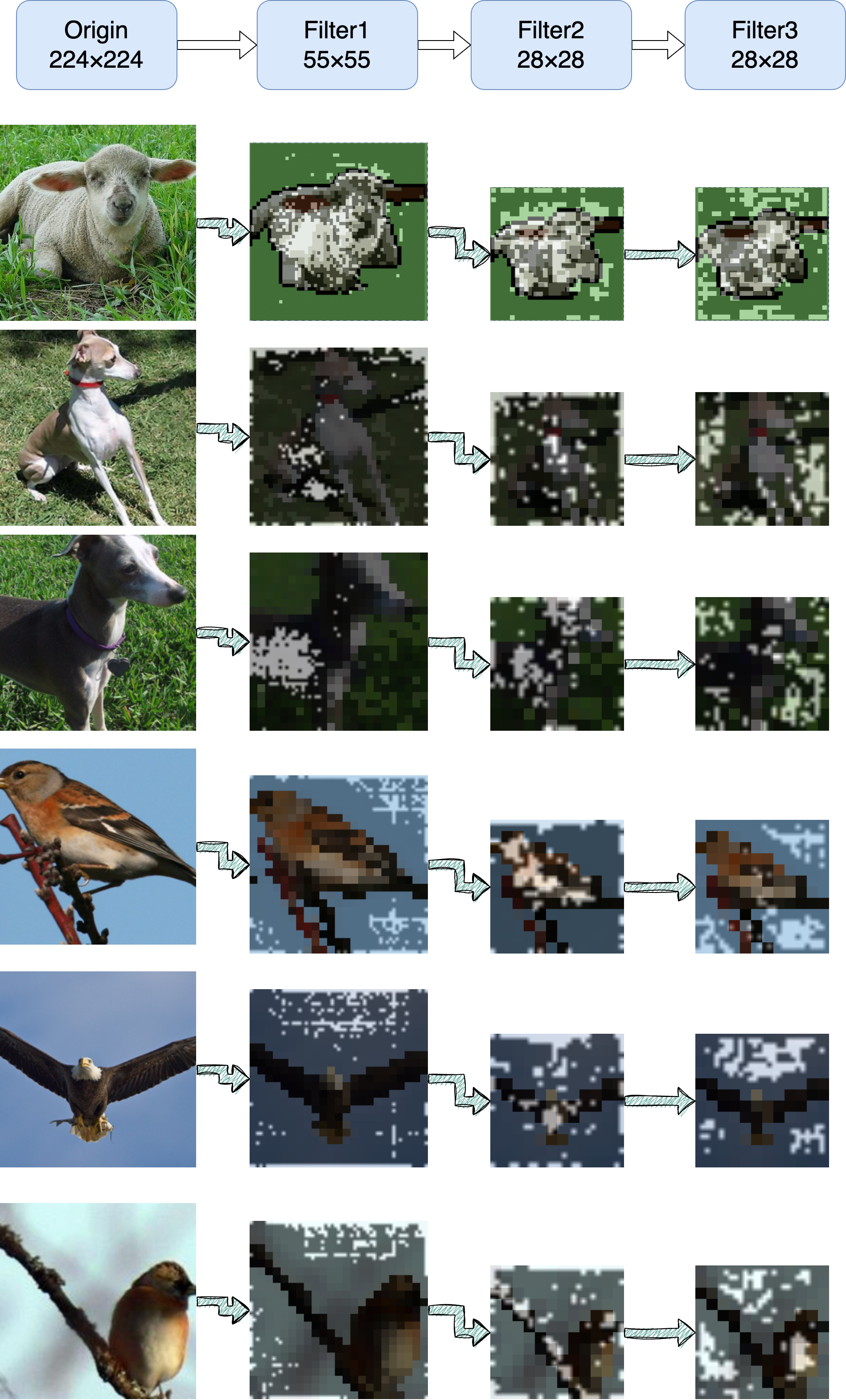}
    \caption{Visualization of filter masks across three Filter Attention layers. The first layer focuses on edges and general features, the second on the main object, and the third on the background. This layered attention mechanism suggests that FilterMobileVit has inherent explainability.}
    \label{fig:explain}
\end{figure}

This collaborative behavior between the layers demonstrates that the Filter Attention mechanism not only selects salient regions for computation but also helps the model understand the image at different granularities. By progressively refining its attention from edges to the main object and then to the background, FilterMobileVit exhibits explainability in its decision-making process, which can be visualized as seen in Fig.~\ref{fig:explain}.

These observations highlight that the model’s filter masks are not arbitrary but rather reflect a structured approach to understanding the image. Each layer contributes a distinct aspect of the image, allowing the model to efficiently balance attention between foreground objects and the broader context. This layered attention structure not only improves performance but also provides a more interpretable understanding of how the model processes visual information, simultaneously validating the correctness of the design.

\section*{Ablation Study: DropoutVIT}
In this study, we introduced a variant of our FilterMobileViT  model called DropOutVIT. The key difference between DropOutVIT and FilterMobileViT  lies in how the model selects pixels for attention computation. While FilterMobileViT  utilizes a CNN-generated filter mask to deterministically select the most salient pixels, DropOutVIT replaces this mechanism with a random sampling approach, where a subset of pixels is randomly selected to pass through the Transformer Encoder during each training step. This can be considered analogous to applying dropout in the attention computation, introducing a degree of stochasticity into the selection process \cite{srivastava2014dropout}.

The rationale behind DropOutVIT is to explore whether random sampling of pixels could lead to similar or improved performance by preventing overfitting to specific features. By introducing randomness, we hypothesize that the model may generalize better, especially during the early training stages when the filter mask in FilterMobileViT  is not yet fully optimized.

\subsubsection{DropOutVIT vs. FilterMobileViT }
\begin{figure}[htbp]
    \centering
    \includegraphics[width=0.5\textwidth]{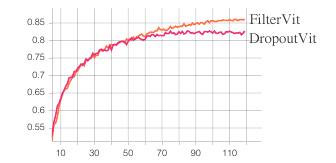}
    \caption{Validation accuracy comparison between DropOutVIT and FilterMobileViT  over the course of training.}
    \label{fig:dropoutvit_val}
\end{figure}

We compared the performance of DropOutVIT and FilterMobileViT  across several training runs (see Fig.~\ref{fig:dropoutvit_val}). Initially, both models exhibited similar performance, with DropOutVIT even outperforming FilterMobileViT  slightly during the early epochs. This may be attributed to the increased diversity in the features selected by the random sampling process, which potentially prevents early overfitting.

However, as training progressed, DropOutVIT appeared to reach a performance plateau around the 50th epoch, whereas FilterMobileViT  continued to improve. One plausible explanation for this is that the CNN in FilterMobileViT  gradually learns to generate more meaningful filter masks, which allows the model to focus on the most critical pixels for attention computation. This refined selection process likely enables FilterMobileViT  to achieve better overall performance, as the model increasingly benefits from targeted attention.

Despite its plateau, DropOutVIT remains a promising variant. The random sampling approach shows potential, especially in scenarios where the deterministic nature of the filter mask may lead to overfitting or reduced generalization. Future work could explore modifications such as adjusting the dropout ratio, experimenting with alternative sampling methods (e.g., Gaussian sampling or central sampling), or applying the dropout mask selectively to certain layers to further optimize DropOutVIT’s performance.

\section{Conclusion}

In this paper, we introduced \textbf{FilterMobileViT}, an efficient and interpretable variant of MobileViT that leverages a Filter Attention mechanism to perform attention-based operations on high-resolution feature maps. By employing a convolutional neural network to generate an Filter mask, our method selectively focuses on the most informative pixels, significantly reducing the computational complexity associated with traditional attention mechanisms.

Experimental results on multiple img-100 subsets of the ImageNet-1K dataset demonstrate that FilterMobileViT achieves a remarkable balance between accuracy and efficiency. Specifically, our model achieves an accuracy of 86.1\% with only 1.89 million parameters, outperforming several baseline models in terms of both performance and computational resource consumption. The inherent interpretability of our model is evidenced by the visualization of filter masks, which highlight the critical regions of the input images that contribute to the model’s predictions.

Furthermore, our ablation study on \textbf{DropoutVIT} reveals that while random pixel selection can offer initial performance gains, the deterministic filtering approach of FilterMobileViT ultimately leads to superior accuracy and convergence. This underscores the effectiveness of our filter mask in guiding the attention mechanism.

Future work will explore extending the Filter Attention mechanism to other transformer-based architectures and investigating alternative methods for generating the filter mask, such as using attention scores or integrating with other interpretability techniques. Additionally, we plan to evaluate our model on larger and more diverse datasets to further validate its generalizability.

In summary, FilterMobileViT represents a significant step towards creating efficient, accurate, and interpretable vision models suitable for deployment in resource-constrained environments, contributing valuable insights to the ongoing development of lightweight neural network architectures.

\bibliographystyle{IEEEtran}
\bibliography{reference}

\begin{thebibliography}{10}
\providecommand{\url}[1]{#1}
\csname url@samestyle\endcsname
\providecommand{\newblock}{\relax}
\providecommand{\bibinfo}[2]{#2}
\providecommand{\BIBentrySTDinterwordspacing}{\spaceskip=0pt\relax}
\providecommand{\BIBentryALTinterwordstretchfactor}{4}
\providecommand{\BIBentryALTinterwordspacing}{\spaceskip=\fontdimen2\font plus
\BIBentryALTinterwordstretchfactor\fontdimen3\font minus \fontdimen4\font\relax}
\providecommand{\BIBforeignlanguage}[2]{{%
\expandafter\ifx\csname l@#1\endcsname\relax
\typeout{** WARNING: IEEEtran.bst: No hyphenation pattern has been}%
\typeout{** loaded for the language `#1'. Using the pattern for}%
\typeout{** the default language instead.}%
\else
\language=\csname l@#1\endcsname
\fi
#2}}
\providecommand{\BIBdecl}{\relax}
\BIBdecl

\bibitem{dosovitskiy2020image}
A.~Dosovitskiy \emph{et~al.}, ``An image is worth 16x16 words: Transformers for image recognition at scale,'' arXiv preprint arXiv:2010.11929, 2020.

\bibitem{vaswani2017attention}
A.~Vaswani, N.~Shazeer, N.~Parmar, J.~Uszkoreit, L.~Jones, A.~N. Gomez, {\L}.~Kaiser, and I.~Polosukhin, ``Attention is all you need,'' in \emph{Advances in Neural Information Processing Systems (NeurIPS)}, 2017.

\bibitem{mehta2021mobilevit}
S.~Mehta and M.~Rastegari, ``Mobilevit: Light-weight, general-purpose, and mobile-friendly vision transformer,'' arXiv preprint arXiv:2110.02178, 2021.

\bibitem{beltagy2020longformer}
I.~Beltagy, M.~E. Peters, and A.~Cohan, ``Longformer: The long-document transformer,'' arXiv preprint arXiv:2004.05150, 2020.

\bibitem{choromanski2020rethinking}
K.~Choromanski, V.~Likhosherstov, D.~Dohan, X.~Song, A.~Gane, T.~Sarlos, P.~Hawkins, J.~Davis, A.~Mohiuddin, {\L}.~Kaiser \emph{et~al.}, ``Rethinking attention with performers,'' arXiv preprint arXiv:2009.14794, 2020.

\bibitem{wang2020linformer}
S.~Wang, B.~Z. Li, M.~Khabsa, H.~Fang, and H.~Ma, ``Linformer: Self-attention with linear complexity,'' \emph{arXiv preprint arXiv:2006.04768}, 2020.

\bibitem{reformer}
N.~Kitaev, {\L}.~Kaiser, and A.~Levskaya, ``Reformer: The efficient transformer,'' \emph{arXiv preprint arXiv:2001.04451}, 2020.

\bibitem{sandler2018mobilenetv2}
M.~Sandler, A.~Howard, M.~Zhu, A.~Zhmoginov, and L.-C. Chen, ``Mobilenetv2: Inverted residuals and linear bottlenecks,'' in \emph{Proceedings of the IEEE conference on computer vision and pattern recognition}, 2018, pp. 4510--4520.

\bibitem{howard2019searching}
A.~Howard, M.~Sandler, G.~Chu, L.-C. Chen, B.~Chen, M.~Tan, W.~Wang, Y.~Zhu, R.~Pang, V.~Vasudevan \emph{et~al.}, ``Searching for mobilenetv3,'' in \emph{Proceedings of the IEEE/CVF international conference on computer vision}, 2019, pp. 1314--1324.

\bibitem{tan2019efficientnet}
M.~Tan, ``Efficientnet: Rethinking model scaling for convolutional neural networks,'' \emph{arXiv preprint arXiv:1905.11946}, 2019.

\bibitem{han2020ghostnet}
K.~Han, Y.~Wang, Q.~Tian, J.~Guo, C.~Xu, and C.~Xu, ``Ghostnet: More features from cheap operations,'' in \emph{Proceedings of the IEEE/CVF conference on computer vision and pattern recognition}, 2020, pp. 1580--1589.

\bibitem{cui2021pp}
C.~Cui, T.~Gao, S.~Wei, Y.~Du, R.~Guo, S.~Dong, B.~Lu, Y.~Zhou, X.~Lv, Q.~Liu \emph{et~al.}, ``Pp-lcnet: A lightweight cpu convolutional neural network,'' \emph{arXiv preprint arXiv:2109.15099}, 2021.

\bibitem{xie2022vit}
W.~Xie, X.-H. Li, C.~C. Cao, and N.~L. Zhang, ``Vit-cx: Causal explanation of vision transformers,'' \emph{arXiv preprint arXiv:2211.03064}, 2022.

\bibitem{russakovsky2015imagenet}
O.~Russakovsky, J.~Deng, H.~Su, J.~Krause, S.~Satheesh, S.~Ma, Z.~Huang, A.~Karpathy, A.~Khosla, M.~Bernstein \emph{et~al.}, ``Imagenet large scale visual recognition challenge,'' \emph{International journal of computer vision}, vol. 115, pp. 211--252, 2015.

\bibitem{muller2021trivialaugment}
S.~G. M{\"u}ller and F.~Hutter, ``Trivialaugment: Tuning-free yet state-of-the-art data augmentation,'' in \emph{Proceedings of the IEEE/CVF international conference on computer vision}, 2021, pp. 774--782.

\bibitem{loshchilov2016sgdr}
I.~Loshchilov and F.~Hutter, ``Sgdr: Stochastic gradient descent with warm restarts,'' \emph{arXiv preprint arXiv:1608.03983}, 2016.

\bibitem{loshchilov2017decoupled}
I.~Loshchilov, ``Decoupled weight decay regularization,'' \emph{arXiv preprint arXiv:1711.05101}, 2017.

\bibitem{tinyvit2022}
B.~Wu, F.~Xia, X.~Huang, Y.~Lin, H.~Wang, P.~Zhu, V.~Sreenivas, A.~Neelakantan, W.~Wei, Z.~Zhang \emph{et~al.}, ``Tinyvit: Fast pretraining distillation for small vision transformers,'' \emph{arXiv preprint arXiv:2207.10666}, 2022.

\bibitem{graham2021levit}
B.~Graham, A.~El-Nouby, H.~Touvron, P.~Stock, A.~Joulin, H.~J{\'e}gou, and M.~Douze, ``Levit: a vision transformer in convnet's clothing for faster inference,'' in \emph{Proceedings of the IEEE/CVF International Conference on Computer Vision}, 2021, pp. 12\,259--12\,269.

\bibitem{han2020model}
K.~Han, Y.~Wang, Q.~Zhang, C.~Xu, and C.~Xu, ``Model rubik's cube: Twisting resolution, depth and width for tinynets,'' \emph{Advances in Neural Information Processing Systems}, vol.~33, pp. 190--202, 2020.

\bibitem{hendricks2021attention}
L.~A. Hendricks, R.~Hu, T.~Darrell, and R.~Girshick, ``Attention rollout: Explaining vision transformer,'' in \emph{Proceedings of the IEEE/CVF International Conference on Computer Vision (ICCV)}, 2021.

\bibitem{kumar2021visualizing}
S.~Kumar, A.~Srinivasan, and D.~Kumar, ``Visualizing and understanding vision transformers,'' in \emph{Proceedings of the IEEE/CVF International Conference on Computer Vision (ICCV)}, 2021.

\bibitem{selvaraju2017grad}
R.~R. Selvaraju, M.~Cogswell, A.~Das, R.~Vedantam, D.~Parikh, and D.~Batra, ``Grad-cam: Visual explanations from deep networks via gradient-based localization,'' in \emph{Proceedings of the IEEE international conference on computer vision}, 2017, pp. 618--626.

\bibitem{srivastava2014dropout}
N.~Srivastava, G.~Hinton, A.~Krizhevsky, I.~Sutskever, and R.~Salakhutdinov, ``Dropout: a simple way to prevent neural networks from overfitting,'' \emph{The journal of machine learning research}, vol.~15, no.~1, pp. 1929--1958, 2014.

\end{thebibliography}
\section*{Appendix}

In this section, we provide a comprehensive comparison of model performance across different image subsets. Table \ref{table:accuracy} summarizes the validation accuracy achieved by each model on the 2nd to 5th subsets of the img-100 dataset, highlighting the effectiveness of our proposed model (“ours”) in comparison to baseline models.

\begin{table}[h]
\centering
\begin{tabularx}{\columnwidth}{|l|X|X|X|X|}
\hline
\textbf{Model} & \textbf{2nd subset} & \textbf{3rd subset} & \textbf{4th subset} & \textbf{5th subset} \\
\hline
MobileNetV2 & 0.8344 & 0.8292 & 0.833 & 0.8226 \\
MobileNetV3\_L & 0.8328 & 0.827 & 0.8362 & 0.8298 \\
MobileNetV3\_S & 0.7898 & 0.7816 & 0.7924 & 0.7736 \\
EfficientNet & 0.8396 & 0.8402 & 0.8428 & 0.8408 \\
Tiny-ViT-11M & 0.877 & 0.8682 & 0.8772 & 0.8696 \\
LeViT & 0.7838 & 0.7812 & 0.7858 & 0.7738 \\
GhostNet & 0.8416 & 0.8368 & 0.837 & 0.8328 \\
TinyNet & 0.8476 & 0.8418 & 0.848 & 0.8398 \\
LCNet & 0.8158 & 0.7982 & 0.7907 & 0.7976 \\
MobileViT & 0.839 & 0.853 & 0.842 & 0.828 \\
FilterMobileViT  (ours) & 0.8578 & 0.8608 & 0.859 & 0.8458 \\
\hline
\end{tabularx}
\caption{Validation accuracy for each model across img-100 subsets (2nd to 5th).}
\label{table:accuracy}
\end{table}

Figures 7 to 14 illustrate the validation accuracy and loss curves for various models across the selected subsets. Each figure set (left and right columns) presents the performance progression over training epochs, with our proposed model consistently outperforming the baselines. Specifically:

The left-column figures (Figures 7, 9, 11, and 13) show a steady increase in accuracy over epochs. Our model generally converges faster and achieves higher final accuracy compared to the baseline models.
The right-column figures (Figures 8, 10, 12, and 14) display a consistent decrease in validation loss over epochs. Notably, our model attains a lower final loss, reflecting improved learning efficiency and generalization.

These results collectively demonstrate the robustness of our proposed method, which not only achieves higher accuracy but also converges more quickly and with lower validation loss, showcasing its potential across different subsets.

\begin{figure}[h]
    \centering
    \includegraphics[width=0.5\textwidth]{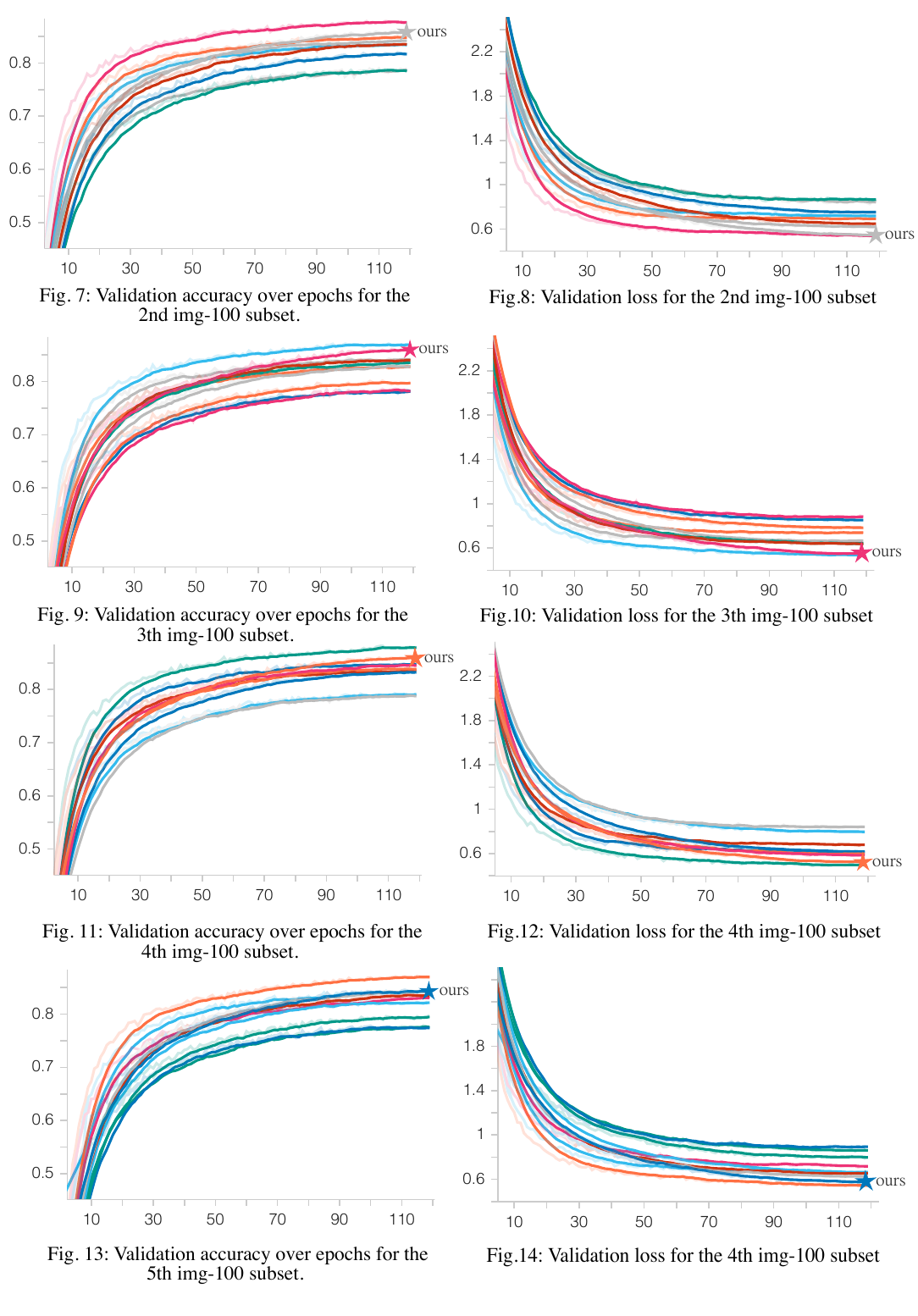}
\end{figure}

\end{document}